**Constructing Evacuation Evolution Patterns and Decisions Using Mobile Device Location Data: A Case Study of Hurricane Irma**

Aref Darzi, Graduate Research Assistant
Department of Civil and Environmental Engineering, Maryland Transportation Institute
1124 Glenn Martin Hall, University of Maryland
College Park, MD 20742
Phone: (240)760-0845, Email: adarzi@umd.edu
ORCID: 0000-0003-2558-5570

Vanessa Frias-Martinez, Associate Professor
Department of Information Studies
4144 Iribe Center, University of Maryland
College Park, MD 20742
Phone: (301)405-2032, Email: vfrias@umd.edu
ORCID: 0000-0001-5114-7633

Sepehr Ghader, Research Scientist
Department of Civil and Environmental Engineering, Maryland Transportation Institute
1124 Glenn Martin Hall, University of Maryland
College Park, MD 20742
Phone: (703)638-4046, Email: sghader@umd.edu
ORCID: 0000-0003-1938-7914

Hannah Younes, Graduate Research Assistant
Department of Geographical Sciences
Maryland Transportation Institute
University of Maryland
College Park, MD 20742
Phone: (410)292-4989, Email: hyounes@umd.edu
ORCID: 0000-0003-4795-3565

Lei Zhang, Herbert Rabin Distinguished Professor (Corresponding Author)
Director, Maryland Transportation Institute
Department of Civil and Environmental Engineering
1173 Glenn Martin Hall, University of Maryland
College Park, MD 20742
Phone: (301)405-2881, Email: lei@umd.edu
ORCID: 0000-0002-3372-6321

## ABSTRACT

Understanding individuals' behavior during hurricane evacuation is of paramount importance for local, state, and government agencies hoping to be prepared for natural disasters. Complexities involved with human decision-making procedures and lack of data for such disasters are the main reasons that make hurricane evacuation studies challenging. In this paper, we utilized a large mobile phone Location-Based Services (LBS) data to construct the evacuation pattern during the landfall of Hurricane Irma. By employing our proposed framework on more than 11 billion mobile phone location sightings, we were able to capture the evacuation decision of 807,623 smartphone users who were living within the state of Florida. We studied users' evacuation decisions, departure and reentry date distribution, and destination choice. In addition to these decisions, we empirically examined the influence of evacuation order and low-lying residential areas on individuals' evacuation decisions. Our analysis revealed that 57.92% of people living in mandatory evacuation zones evacuated their residences while this ratio was 32.98% and 33.68% for people living in areas with no evacuation order and voluntary evacuation order, respectively. Moreover, our analysis revealed the importance of the individuals' mobility behavior in modeling the evacuation decision choice. Historical mobility behavior information such as number of trips taken by each individual and the spatial area covered by individuals' location trajectory estimated significant in our choice model and improve the overall accuracy of the model significantly.

# 1. INTRODUCTION

Three of the top five costliest U.S. hurricanes occurred in 2017, making the 2017 Atlantic hurricane season the costliest on the record[1,2]. These three hurricanes, Hurricane Harvey, Maria, and Irma impacted the lives of millions of people in several states in the United States. The complexity of managing the evacuation, allocating accessible transportation for different society groups, and providing safe shelters for evacuees are among the key challenges that agencies face during such disasters.

In September 2017, Hurricane Irma prompted officials to issue one of the largest evacuation orders in U.S. history. Over six million people were ordered to evacuate their residences due to Irma's landfall in Florida, Georgia, and South Carolina. Mandatory and voluntary evacuation orders were issued before the landfall of the storm, in both Atlantic and Gulf coasts. 84 deaths were reported only in the state of Florida due to either direct effect of Hurricane Irma such as drowning or indirect causes such as vehicle accidents during the evacuation[3]. The immense scale of Irma hurricane and the dependence of the evacuation management on how people behave during these disasters highlighted the importance of studying the evacuation patterns of the people in such big disasters.

To study evacuees' behavior, post-hurricane surveys are traditionally used to collect information regarding various evacuation decisions (i.e., evacuating or not, departure time of the evacuation, destination choice, primary travel mode used for the evacuation, route choice, and reentry time decision) [2,4,5]. Although this type of survey is rich in terms of recording evacuees' decisions and revealing their preferences during the disaster, such surveys are costly, implemented for a small number of respondents, time-consuming, and not capable of providing real-time information.

With the increasing availability and popularity of big-data, new approaches are now available to tackle old problems. Billions of location data points are being passively collected from mobile devices, producing a blueprint of people's movement patterns. Mobile device location data sources include cell phone call detail record (CDR), GPS location data from in-vehicle GPS devices, and smartphone app location data, also known as Location-Based Services (LBS) data. Smartphone apps often record the phone's location for various services, using GPS, Wi-Fi, Bluetooth, etc. Once privacy concerns are addressed, one can refer to passively-collected location data to observe the movement pattern of millions of people before, during, and after any event such as a hurricane. Robinson et al. (2017) identified two main challenges in modeling disaster evacuation; the first is the complexity of human behavior and the second is data deficiency for traffic conditions and household decisions [6]. Both issues can be resolved to some extent by passively-collected location data. Such data does not include detailed individual-level information, but with its significant sample size, it can reveal valuable information and answer many critical questions. Besides the larger sample, passively-collected data has other advantages over traditional surveys. The first is related to a phenomenon known as the observer effect, which describes that individuals may modify their behavior when being observed or studied[7]. Passively collected data capture the normal behavior of subjects, free of any study-related observer error. The second is related to typical survey design errors such as sampling error, measurement error, and response error [8] and survey response biases [9]. Even though passively collected data may have their own biases (such as bias toward higher income or younger individuals) and errors (such as inaccurate sightings), they record the actual behavior, not recalled or stated behavior. The third is related to their ease of availability. Even though passively-collected data may sometimes be costly, they can be available almost in real-time without any major effort for collection. The fourth is specific to disaster-related surveys. Surveying individuals about a traumatic event such as a hurricane may sometimes be undesired for the respondents. Passive data collection does not put any emotional burden on the respondents.

Some previous studies attempted to analyze evacuee's behavior using big-data from social media platforms (e.g. Twitter, Facebook) [10,11]. The social media data such as the data from tweets are usually geo-coded, which provides low-frequent data with some contexts. The higher frequency of location sighting, higher penetration rate, and smaller demographic bias are the key advantages of the passively-collected location data over the social media data. Among the mobile device location data sources, LBS is becoming more popular because of its high penetration rate, frequent and precise sightings (in comparison with CDR), and multi-modality (in comparison with in-vehicle GPS). LBS dataset has been recently used for studying behavior before, during, and after disasters [12-14]. The recent studies highlight the value of LBS data in studying community-level evacuation behavior.

This study is the first to take advantage of mobile location big-data in modeling individual-level evacuation behavior during disasters. We employed LBS data collected from smartphone apps to understand the evacuation patterns and individual-level evacuation decisions during Hurricane Irma. Our LBS dataset consists of anonymized location data for more than 25% of the entire U.S. smartphone users. In this paper, we analyzed two months of LBS data for devices within the state of Florida to capture the actual evacuation pattern of a sample of Florida residents. By identifying the home location of each anonymized user and analyzing their movement trajectory, we were able to analyze the evacuation behavior of 807,623 anonymized smartphone users in Florida. Using machine learning methods and computational algorithms, we were able to add more context to the passively collected location data by imputing home locations and identifying departure and reentry dates of evacuees. This study highlights the potential of passively collected location data in constructing travel behavior, especially evacuation-related behavior. Our findings show great accordance with results from previous studies, highlighting the validity of our efforts. The findings of this paper provide empirical evidence for decision-makers on how people evacuate during big disaster events such as Hurricane Irma.

In addition to constructing the evacuation pattern, we further examined the importance of the individuals' mobility behavior in their decision-making procedure. Our statistical model revealed that people who made more trips during the pre-disaster condition and covered a larger spatial area in their daily movements in pre-disaster time are more likely to evacuate their home location during the disaster.

The rest of this paper is organized as follows: the next section summarizes previous studies related to understanding evacuation behavior, followed by a description of the data used in this study. Next, the methodology is discussed in detail. Evacuation pattern was constructed in the next part followed by the statistical model for evacuation decision choice. The paper concludes with a summary of major findings.

## 2. RELATED STUDIES

There is a wide range of research studies focused on disasters such as hurricanes. Most related to our study are those focusing on evacuation behavior. Several studies reviewed the literature of evacuation behavior [15], evacuation modeling [16], and evacuation practices [17].

Many of the studies are focused on a specific disaster or set of disasters to analyze the important factors in evacuation behavior, evaluate the disaster planning and preparation, or assess disaster management and logistics. A recent example is Collier et al. (2019) studying major transportation and logistics issues and lessons learned for two major hurricanes in the U.S., Hurricane Katrina and Hurricane Harvey. They provided recommendations for future hurricanes in terms of evacuation planning, information provision, infrastructure management, and disaster preparation [18]. Several disaster planning and management studies relied on simulation models [19-25]. A recent example is Feng and Lin (2019), which used a hurricane-prediction demand generation model in a fast agent-based modeling framework calibrated with traffic observations to study the case of evacuation in Hurricane Irma [22].

The majority of evacuation behavior studies rely on surveys [2,26-30]. These surveys ask the impacted individuals about their behavior their choices, and their opinions. For instance, Kontou et al. (2017) collected telephone survey data from commuters affected by Hurricane Sandy and estimated a hazard-based model to identify the parameters that affect the duration of commute behavior changes [29]. Wong et al. (2018) collected an online survey from individuals impacted by Hurricane Irma and studied their evacuation behavior [2]. They offered descriptive statistics and discrete choice models for various choices made during disaster and variables affecting the choices. Wong et al. (2019) used the same survey to study if sharing economy could improve the transportation and sheltering resources for the vulnerable population and improve the equity in evacuation resources [30].

As discussed in the introduction, more recent studies are taking advantage of big-data for evacuation behavior, for its larger sample size, ease of collection, and empirically observed information. The dominant datasets in these studies are in the form of social media data. Kumar and Ukkusuri (2018) utilized geo-tagged tweets from New York City at the time of Hurricane Sandy to study the evacuation behavior of affected residents [31]. Their study showed a strong relationship between social connectivity and the decision to evacuate. Roy and Hasan (2019) collected Twitter data related to Hurricane Irma and developed a hidden Markov framework to model dynamics of hurricane evacuation and infer evacuation decisions [11]. Wang and Taylor (2014) also used Twitter data for Hurricane Sandy to study mobility patterns during hurricane and found that the movement patterns under steady-state and perturbed state are highly correlated [10].

We believe the passively-collected mobile location data have the advantage of higher frequency of location sighting, higher penetration rate, and smaller demographic bias in comparison with the social media data; however, applications of such data in evacuation studies are very limited [12,13]. Yabe et al. (2020) collected LBS data for five disastrous events (1.9 Million devices in total) to study recovery patterns at macroscopic population level and showed similarity in recovery patterns of these events despite differences in population characteristics [13]. Yabe et al (2020) used LBS data for 1.7 million devices in Florida to study the effect of income in evacuation and re-entry behavior related to Hurricane Irma and showed significant income inequality [12]. This paper continues this line of research and utilizes LBS data for evacuation behavior. We use LBS data not only to understand the full picture of the behavior before, during, and after disastrous events but also to model individual-level evacuation decisions based on the historical mobility patterns and socio-demographic information. We believe an understanding of individual-level evacuation behavior can help design effective policies that can save lives and reduce the tolls.

## 3. DATA

### 3.1. Location-Based Services Data

The primary dataset used in this study is a location-based services dataset of anonymized smartphone devices for the entire United States gathered by a location intelligence and measurement company Cuebiq. Information in this dataset was recorded passively through mobile phone apps. Each observation includes timestamp of the observation in Unix epoch time format, an anonymized hashed identification number (ID), ID type that represents the device operating system (OS), latitude and longitude coordinates in decimal version, location accuracy associated with each data point in meters, and time zone offset of the position of the device. A synthetic sample of data is given in **Table 1** to demonstrate the raw data. Data presented in **Table 1** is modified in ordered to preserve privacy.

**Table 1. A synthetic sample of LBS data**

| Timestamp | Device ID | Device Type | Latitude | Longitude | Location Accuracy (m) | Time Zone Offset |
|---|---|---|---|---|---|---|
| 1504068337 | e07941996a2ffd303021914e0c12gcf | 1 | 28.43023 | -81.60654 | 5 | -14400 |
| 1504068342 | e07941996a2ffd303021914e0c12gcf | 1 | 28.43038 | -81.60531 | 25 | -14400 |
| 1504068351 | e07941996a2ffd303021914e0c12gcf | 1 | 28.43029 | -81.60427 | 5 | -14400 |
| 1504068360 | e07941996a2ffd303021914e0c12gcf | 1 | 28.43058 | -81.60463 | 100 | -14400 |
| 1504068369 | e07941996a2ffd303021914e0c12gcf | 1 | 28.43139 | -81.60374 | 5 | -14400 |

Based on the meteorological history, Irma developed from a tropical wave near Cape Verde on August 30 and quickly intensified into a Category 3 hurricane by August 31 due to the climate condition. On September 4, the storm kept intensifying, making it a Category 5 hurricane. Therefore, based on the timeline of Hurricane Irma's evolution, we chose the month of August to identify the home location of the users within the state of Florida, as we assume that users' behavior had not been impacted by the news of Hurricane Irma yet. Furthermore, to understand the evacuation pattern of the residents in Florida, the data from the entire month of September were employed.

### 3.2. Evacuation Zone Data

In addition to the location data, gathering information regarding evacuation order evolution was necessary to understand the evacuees' behavior. The Florida Division of Emergency Management provided the spatial polygon of evacuation zones for the counties with defined evacuation zones [32]. However, for the information regarding evacuation orders by county and zones, no single source provided comprehensive details. The webpage of the former Florida governor, Rick Scott, had one of the most complete information regarding the issuance of evacuation orders as of 9/9/2017 [33]. However, several counties, particularly in the north of Florida, issued evacuation orders on 9/10/2017. Also, many counties upgraded from voluntary to mandatory evacuation orders on or after 9/9/2017. Therefore, we looked at several various sources and compiled the data for each evacuation zone. The final Florida map by evacuation order and date during Hurricane Irma is shown in **Figure 1**. Besides the evacuation map, open-source parcel-level information for the entire state of Florida was obtained. The data were gathered by the Florida Department of Revenue, County Property Appraisers, and the University of Florida GeoPlan Center. This layer contains residential home type information that has been used in the parameter selection process for the home location identification algorithm.

Also, to measure the impact of living in low-lying residences on the evacuation decision, the elevation information was obtained from the digital elevation model (DEM) provided by the University of Florida GeoPlan Center for the entire state of Florida.

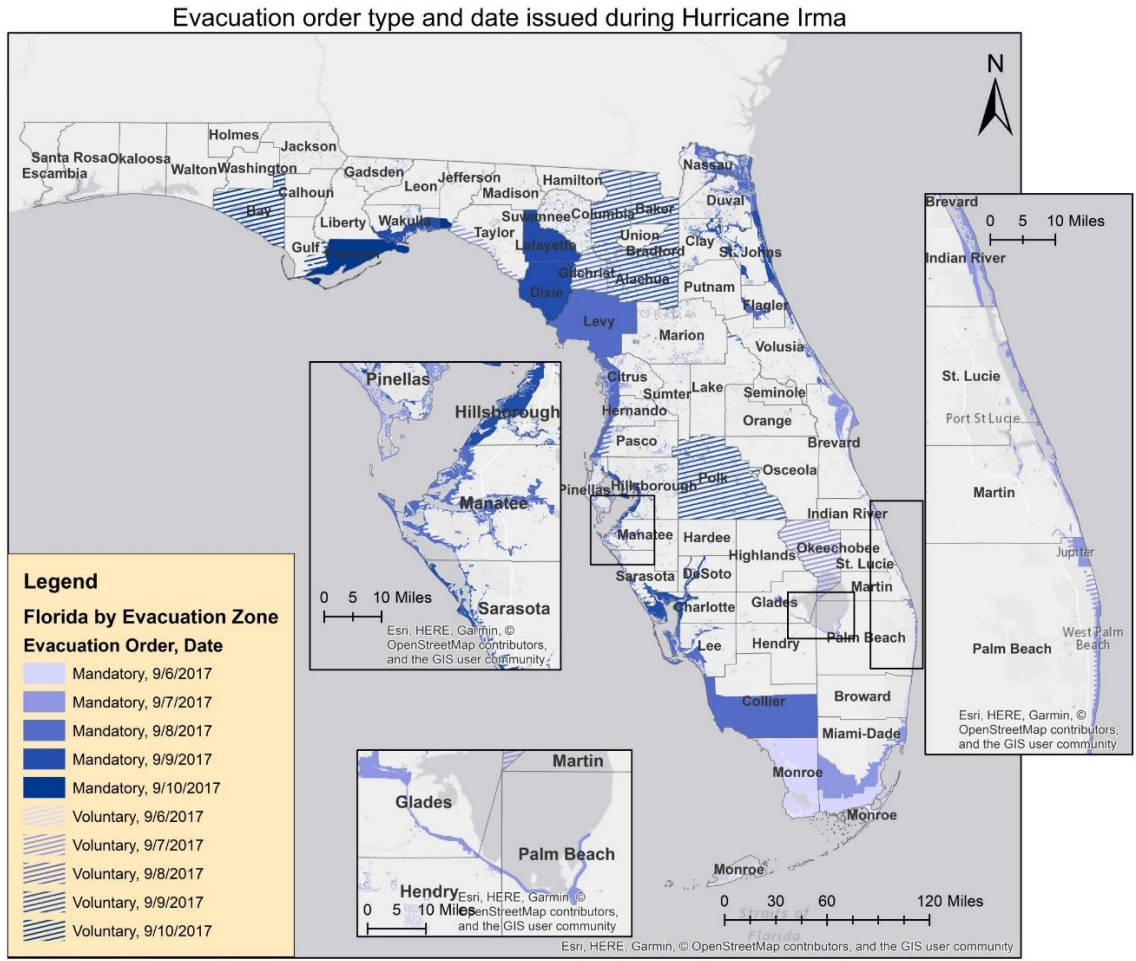

**Figure 1. Florida map by evacuation order and date during Hurricane Irma**

### 3.3. Socio-Demographic Data

In addition to the passively collected location data and evacuation zone information, socio-demographic information such as income, age, and race information was gathered for statistical modeling purposes. We have used the 2017 American Community Survey (ACS) 5-year estimates conducted by the United States Census Bureau to collect socio-demographic information at the census tract level. The census-tract level socio-demographic information was added to devices based on their residential location.

## 4. METHODOLOGY

We describe the three steps of our analysis to capture the evacuation pattern from the LBS data: (1) Identifying the home location of each anonymized users to filter out the devices that are not living within the state of Florida; (2) Detecting devices that evacuated during the Hurricane Irma based on a proposed framework and constructing their evacuation behavior; (3) Calculating mobility metrics such as number of trips and convex hull area for each device daily to develop a descriptive model for evacuation decision based on both mobility and socio-demographic characteristics of the devices. **Figure 2** displays the steps of our methodology.

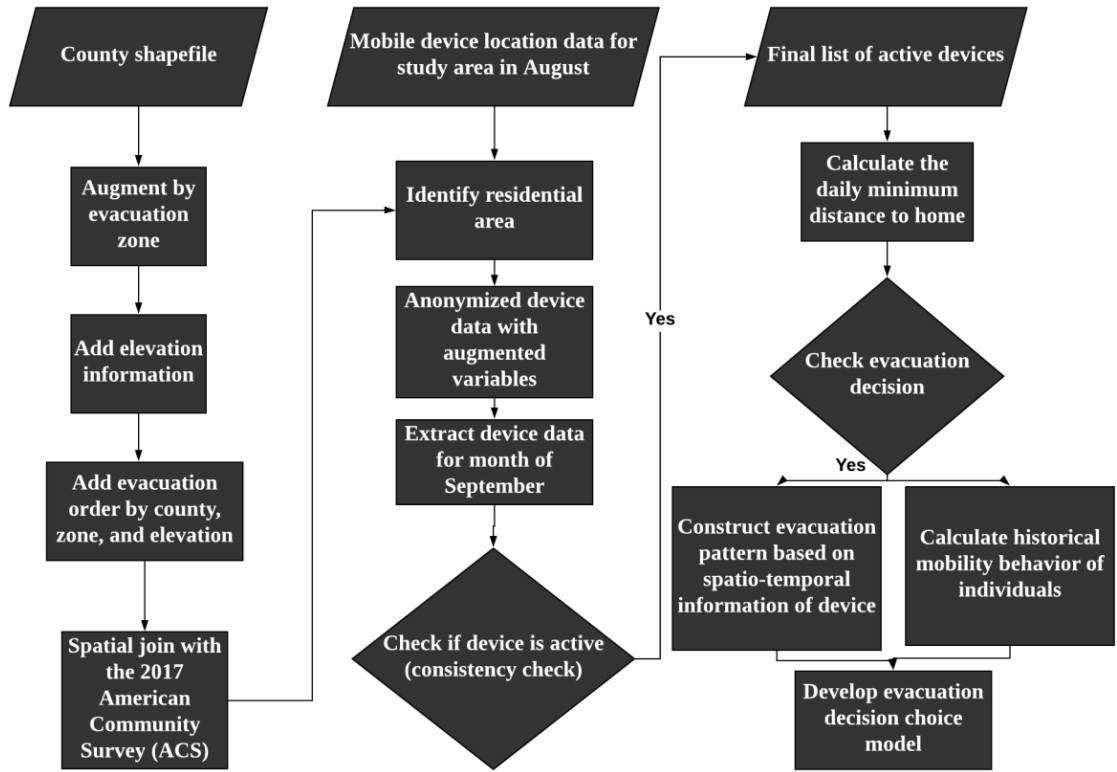

**Figure 2. Methodology flowchart**

## 4.1. Home Location Identification

Identifying the home location of each device was the first step of the analysis. Based on the assumption that people were not impacted by Hurricane Irma during August, thirty-one days of data in August were analyzed to achieve a robust estimation of the home location.

2,132,776 devices were observed at least once in the study area within August. The next step was to identify the home location for all the extracted users by analyzing the 6,210,853,449 sightings.

Many studies in the literature discuss how to infer activity locations from the individual traces. Several clustering algorithms including distance-based, agglomerative, model-based, incremental, and density-based clustering have been implemented on various types of passively collected location data [34-36].

To cluster the traces of each device, the Density-based spatial clustering of applications with noise (DBSCAN) clustering approach was used. DBSCAN is a clustering algorithm relying on a density-based notion of clusters, designed to discover clusters regardless of their shapes [37]. For identifying the home location we implemented the DBSCAN algorithm on each device's GPS sightings in the time window of 7 PM to 7 AM for the entire month of August. Among all clusters determined by the algorithm, the home location was defined as the center of the cluster with the highest frequency of observation.

## 4.2.   Evacuation Detection

After filtering the devices with the inferred home located within the state of Florida, the sighting data of these devices for the entire month of September were extracted to study the evacuation pattern of the residents of Florida during Hurricane Irma.

First, we did some additional checking on the list of the devices to check the persistency of the device IDs and to make sure that devices are still active. To do so, devices with at least one observation in 1-mile radius of their home location during September were kept for further analysis. This condition can remove devices without any information in September as well as devices that changed their home location or were on a trip to Florida during August. After this check, the identified home location of each device was intersected with the augmented shapefile to specify the corresponding county, evacuation zone, elevation information, and socio-demographic attributes of each device.

The next step was to define evacuation based on the available traces of each device. Evacuation identification method was developed based on the distance of the users' sightings to their inferred home location during Hurricane Irma. For this purpose, we calculated the daily minimum distance between each device's sightings and their identified home from September 1st to 30th. A 1-mile threshold was selected as the evacuation criteria. If users were not observed within a 1-mile radius of their home location during the hurricane study period, we assumed that they evacuated their homes. The former Florida Governor, Rick Scott, declared a state of emergency on September 4, and within the next six days, 57 of the 67 counties issued evacuation orders. Eventually, Hurricane Irma made landfall on Cudjoe Key on September 10 as a Category 4 hurricane and exited Florida into Georgia on September 11, after being significantly weakened. Thus, the period between September 4 and September 12 was chosen as our hurricane study period for determining the evacuation decision of the users.

## 4.3.   Mobility Behavior Pattern

In addition to constructing the evacuation pattern, in this paper, we explored the relationship between individuals' mobility behavior and its impact on their evacuation decisions. In particular, we extracted number of trips and convex hull set information for each device during August based on their location trajectories. As the mobile device location does not provide trip information, we applied our previously developed recursive trip identification algorithm to extract trip information. **Figure 3** shows an overview of the trip identification algorithm. A more detailed description of our trip identification algorithm can be found in our previous work [38].

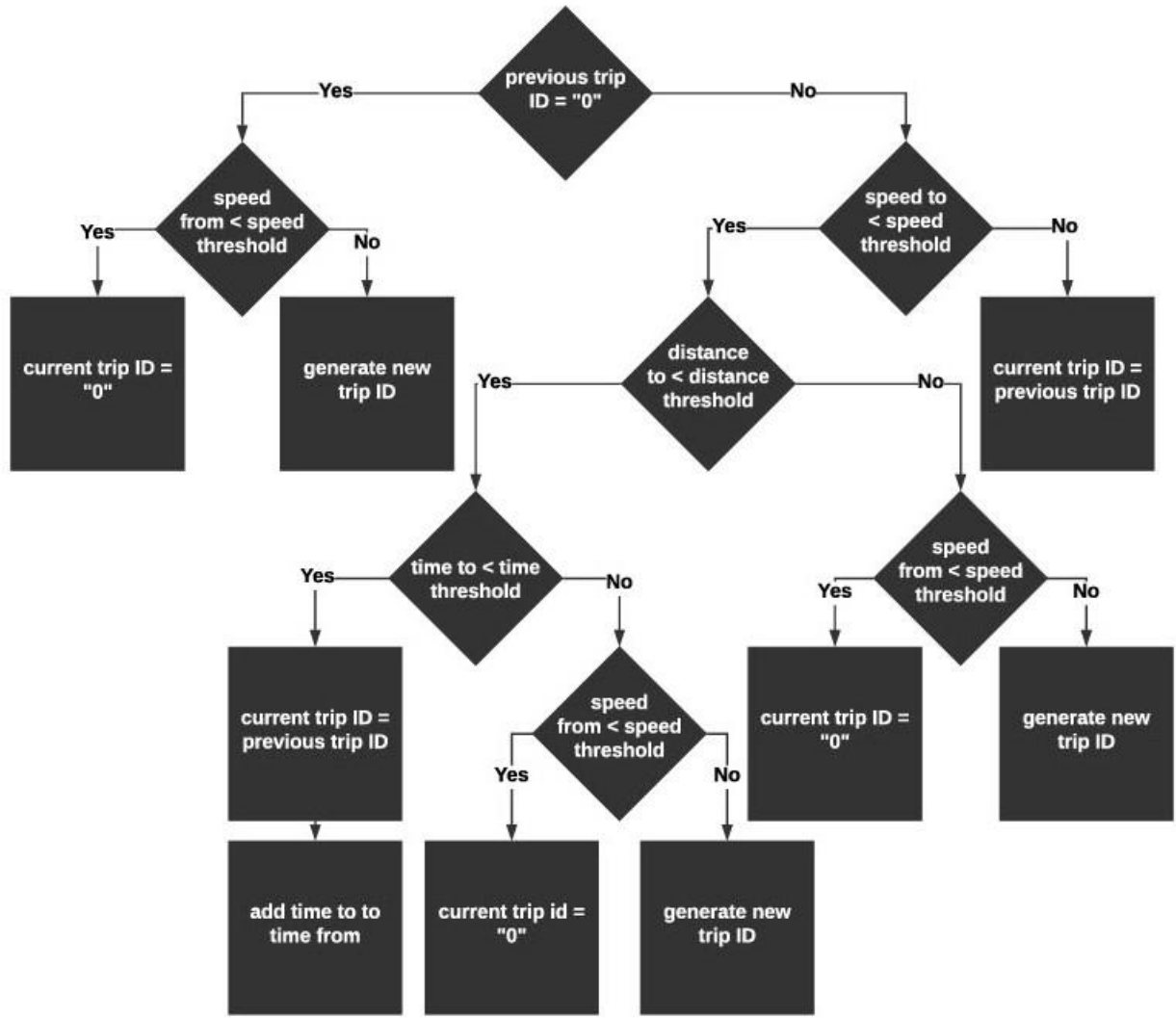

**Figure 3. Recursive trip identification algorithm**

In addition to number of trips, we also calculated the convex hull characteristics of each device daily. Convex hull has been widely used for understanding human mobility behavior based on location trajectories in the literature[39,40]. In this paper, we computed the area and perimeter of the daily convex hull to investigate the relationship between individuals' mobility measures and their evacuation decisions.

## 5. CONSTRUCTING THE EVACUATION PATTERN

In addition to analyzing the evacuation decision, departure date and reentry date are of paramount importance in disaster evacuation management. Therefore, we used the minimum daily distance measures to find out the distribution of the departure and reentry dates. For the devices who evacuated, the latest day before the evacuation in which they were seen in the 1-mile radius of their identified home was chosen as their departure date. Similarly, the earliest day after the evacuation, in which they were seen in the 1-mile radius of their identified home was selected as their re-entry date. Estimating the departure and reentry date enabled us to construct the picture of the evacuation pattern. Moreover, we studied the relationship between the evacuees' departure date and the date that the corresponding evacuation order was issued. Observing how individuals react to the evacuation order helps us better understand the individuals' response to the evacuation orders and plans.

Destination choice is another important decision component. While an increase in short-distance evacuations increases the demand for sheltering resources, it reduces the stress on the transportation network as well as the cost of evacuation. As a result, agencies are becoming increasingly interested in short-distance evacuations. In this paper, we used the maximum value of individuals' minimum distances from their inferred home locations during the evacuation period as the proxy for the distance between their home and the evacuation destination. Also, we empirically examined the impact of living in a low-lying residential area on individuals' evacuation decisions. To have a better understanding of the effects of the low-lying area on the evacuation rate, we controlled for the type of evacuation order in our analysis.

### 5.1. Stay or Evacuate

Implementing the home location identification algorithm discussed above on the 6 billion observations for the devices that were observed in Florida during August, we were able to infer the home location of 1,050,472 devices. Among these devices, 1,002,858 of them resided within the state of Florida. Extracting the information of these devices for September, 5,677,549,347 sightings were filtered from our LBS data and analyzed. The additional checks were conducted to remove inactive devices during September as well as eliminating devices that did not have enough sightings near their home locations. The final list of devices includes 807,623 active devices. The minimum distance from the identified home location was calculated daily for all users. Then the proposed framework for evacuation identification was employed to find out the evacuation decision, departure date, and reentry date of the evacuees. A summary of the rate of evacuation by each evacuation order type is shown in **Table 2**. Based on our results, 57.92% of the people who received mandatory evacuation orders evacuated their homes while this ratio was considerably lower for people who received voluntary evacuation or no evacuation order (33.68% and 32.98%, respectively). These results are in accordance with the results of a telephone poll conducted on October 17, 2017, that showed 57% of people followed the mandatory evacuation order and in general, 33% of the Floridians were evacuated their home [41].

**Table 2. Evacuation decision based on the evacuation order received**

|  | No Evacuation Order | | Voluntary Evacuation Order | | Mandatory Evacuation Order | | Entire State | |
|---|---|---|---|---|---|---|---|---|
|  | Number | Ratio | Number | Ratio | Number | Ratio | Number | Ratio |
| Evacuated | 187285 | 32.98 | 38524 | 33.68 | 72628 | 57.92 | 298437 | 36.95 |
| Not Evacuated | 380547 | 67.02 | 75868 | 66.32 | 52771 | 42.08 | 509186 | 63.05 |
| Total | 567832 | 100 | 114392 | 100 | 125399 | 100 | 807623 | 100 |

### 5.2. Departure and Reentry Date Distribution

Departure and reentry date choices are becoming increasingly important for the emergency and transportation practitioners as well as state and government agencies. We tried to estimate the departure and reentry date distribution by employing the method discussed previously on our LBS dataset. We acknowledge that this approach might have some deficiencies in capturing the actual departure and reentry date accurately for the devices that lost their connections to the network either due to power outage or losing cell network services during and after hurricane landfall.

However, comparing the results from our analysis using LBS data for Hurricane Irma with the conducted survey for the same regions shows consistent patterns between the two outcomes [2]. A summary of the result can be seen in **Figure 4**.

Based on our results, the majority of the evacuations occurred from September 8 to September 9, with September 9 being the peak with 26.27%. Although the majority of evacuations happened in the last three days before Irma's landfall, our results showed that a considerable number of people in our data evacuated their home 5 days or earlier in advance, with 7.04% of people evacuated on September 5 and 10.28% evacuated before September 5. This high rate of early evacuation might be because some counties started to issue evacuation orders as early as September 5. Increased implementation of time-phased evacuation plans can be another reason for our observation. Finally, only 2.13% of the evacuees left their homes after September 10.

On the other hand, reentry date distribution was smoother in comparison to the departure date, with a peak of 24.65% observed on September 11. This was expected since regions do not become livable at once after a disaster. Besides that, agencies do not provide returning plans for the impacted areas. Therefore, people usually decide to re-enter their residence in a way that minimizes any impedance such as traffic. Moreover, our results indicated that about 12.89% of the evacuees returned to their homes on September 10 or earlier. This observation might have happened because of the updates on the hurricane path. Individuals who evacuated earlier might have reached to the conclusion that their residences were no longer at risk. This behavior was observed in the conducted surveys as well [2].

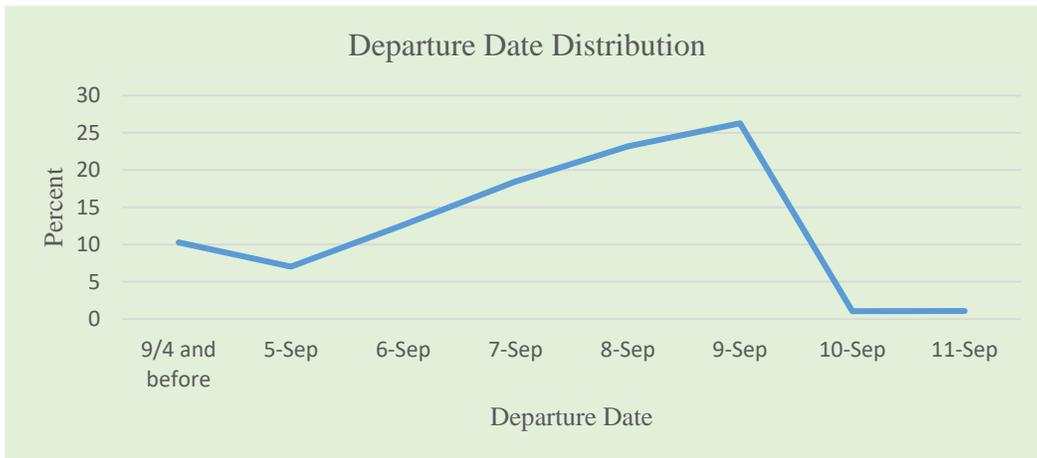

(a) Departure date distribution

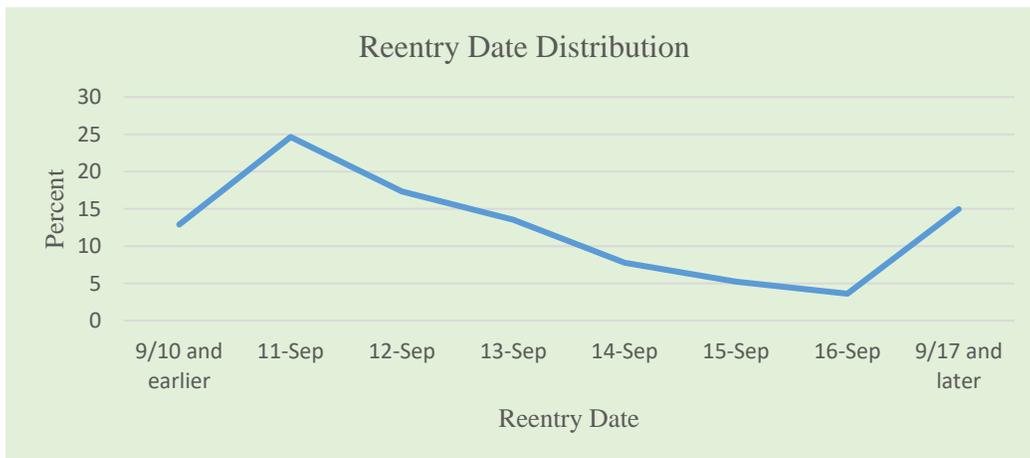

(b) Reentry date distribution

**Figure 4. Departure and reentry date distribution**

To delve more into the departure date distribution, the effect of the corresponding date that evacuation order was issued on the departure date was investigated for all the regions with mandatory or voluntary evacuation orders. The majority of the individuals who received evacuation orders on September 6 departed their homes on September 7 and September 8 while people who received orders on September 7 mostly chose to leave their home between September 7 to September 9. The same trend can be observed for the people who were ordered to evacuate their homes on September 8. 34.53% of them decided to leave their residences on the following day. As it got closer to the landfall date, the impact of the evacuation order date on the individuals' actual departure date decision decreased. The majority of the evacuees who were ordered to evacuate their homes on September 9 and September 10 had already left their home before receiving the evacuation order. **Figure 5** is color-coded by the evacuation order date.

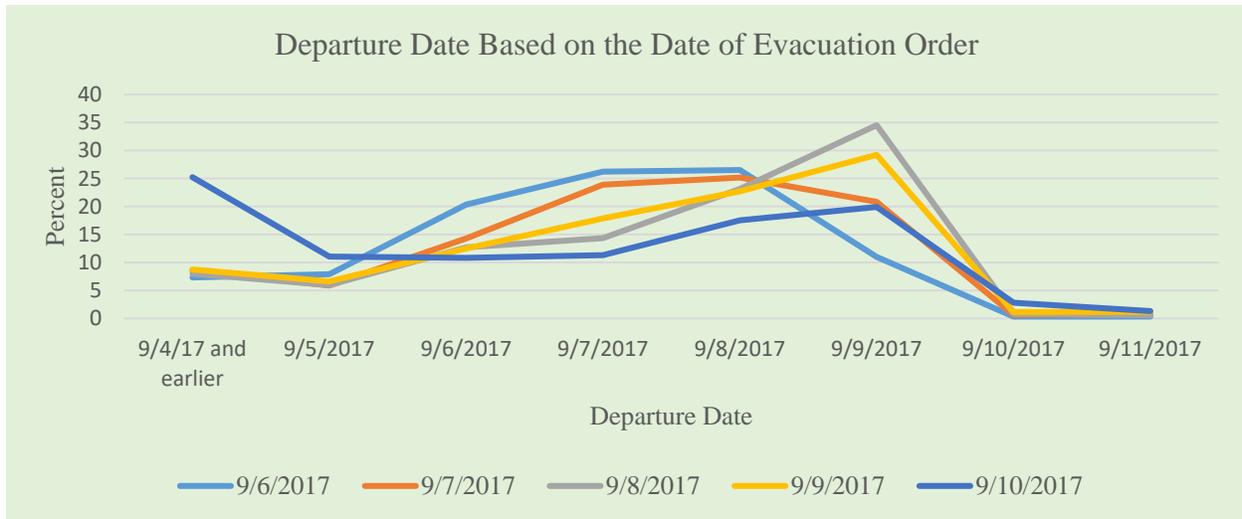

**Figure 5. Relationship between departure date and the date that evacuation order was issued**

## 5.3.  Destination Choice: Distance to Shelter

The overall distribution of distance to shelter followed a similar trend among evacuees who received various evacuation orders. However, on average, evacuees who received mandatory evacuation order traveled to farther locations. The trend is shown in **Figure 6**.  While about 43% of the evacuees who received voluntary evacuation order or no order at all decided to choose shelters within 20 miles radius of their residential locations, 35.47% of evacuees who received mandatory evacuation order stayed within 20 miles radius of their home. A possible reason for this observation might be that people who received mandatory evacuation orders may have not felt safe remaining in their neighborhood regions. The distance distribution also showed that the greater number of evacuees decided either to select shelters within a 20-mile radius of their residential area or to travel to further locations with distances more than 100 miles. It implies that evacuees tend to either choose a close shelter within their neighborhood or travel farther away to get to their perceived safe places.

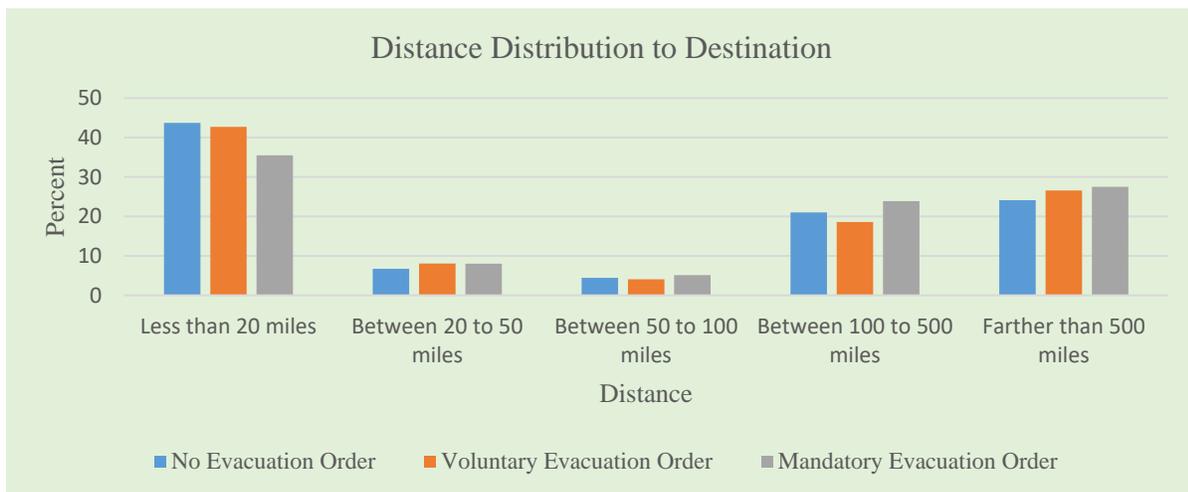

**Figure 6. Distribution of shelter distance to the home locations**

As the overall trend of the evacuation distance was similar for different evacuation group, the spatial distribution of the evacuation distance is also depicted in **Figure 7** for further analysis. Evacuees living near the shores tend to travel to farther destinations. This observation is in line with the expectations as those people perceive higher risk in comparison to the people living in the midland.

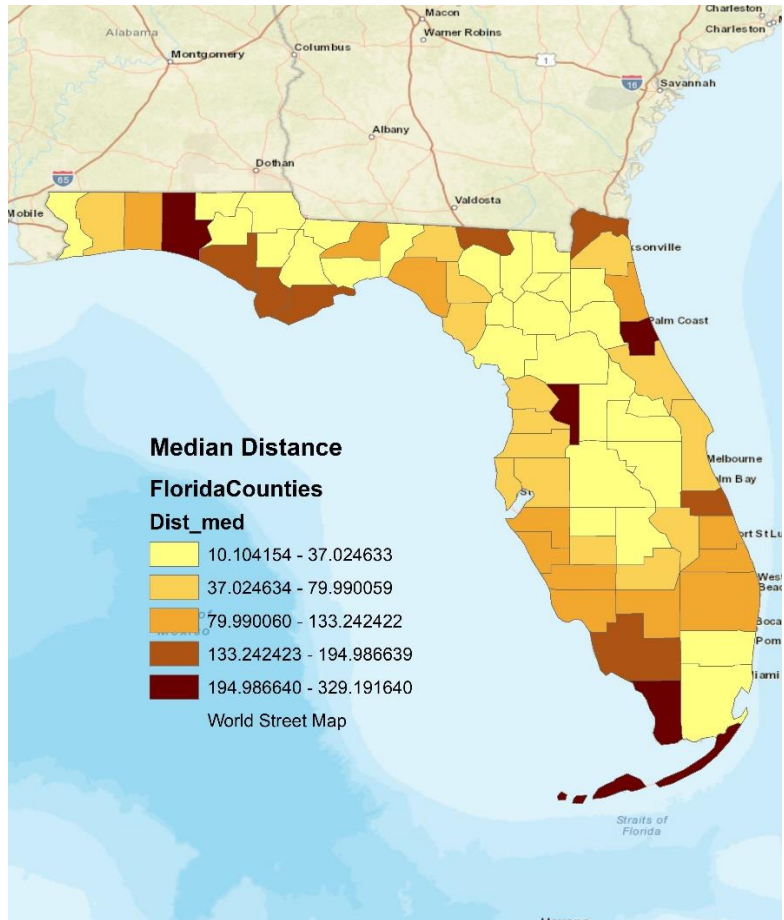

**Figure 7. Median distance to shelter at county level**

### 5.4. Evacuation Duration Distribution

In terms of evacuation duration, as it is shown in **Figure 8**, evacuees who received mandatory evacuation orders had slightly longer evacuation duration. To better understand the spatial trend of evacuation duration, the average evacuation duration at county level is also presented in **Figure 9**. People living in the southern counties of Florida had a longer evacuation duration which might be due to the fact that damages to the properties and infrastructures were more substantial than other regions in Florida.

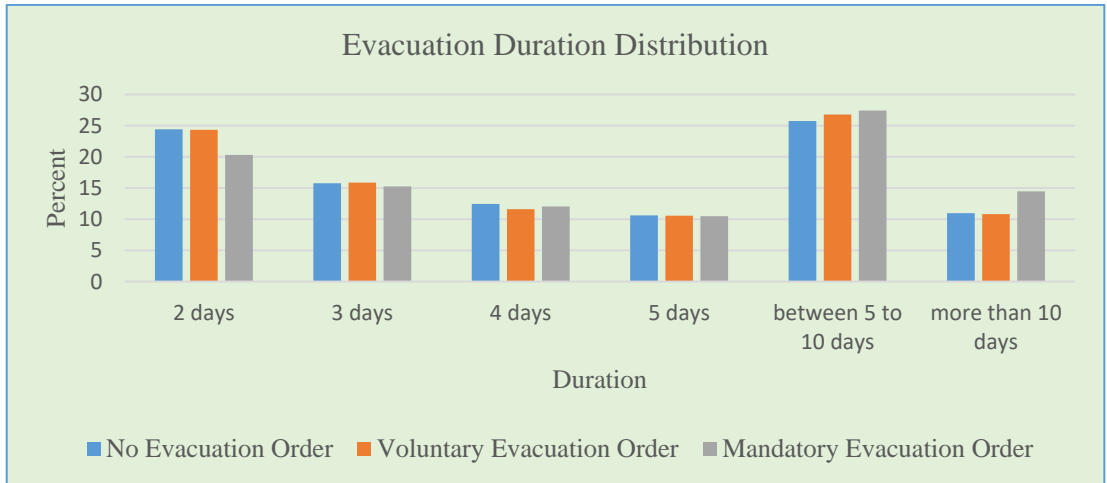

**Figure 8. Evacuation duration distribution among different order groups**

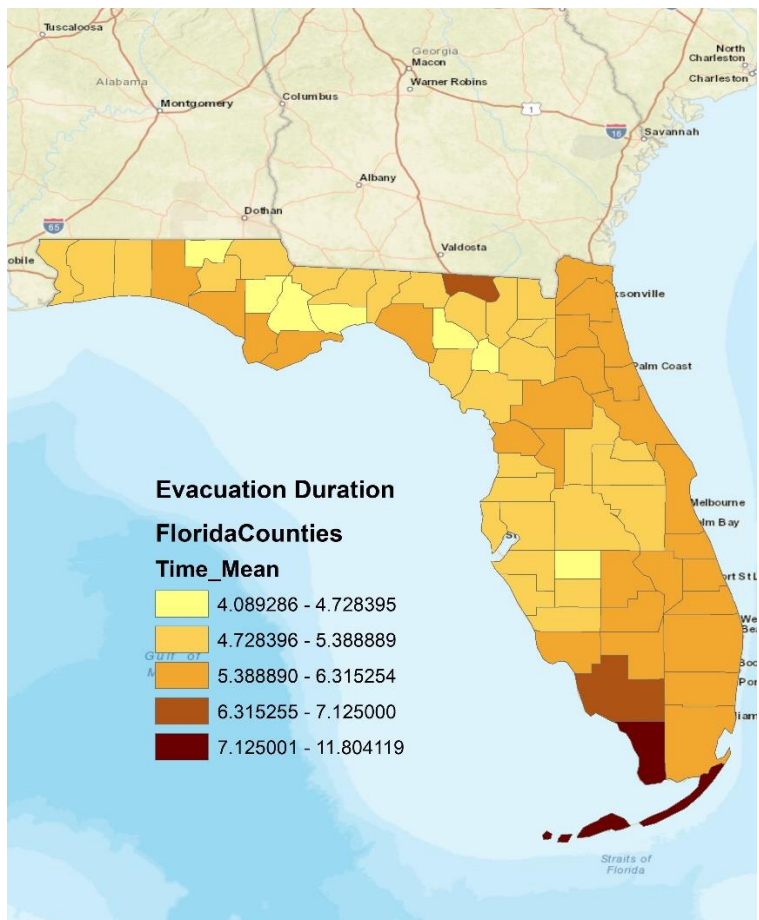

**Figure 9. Average of evacuation duration at the county level**

## 5.5.  Impact of Low-Lying Residential Area

We investigated the impact of low-lying residential areas on individuals' evacuation decisions. As there is no solid definition for low-lying areas, we categorized individuals into three classes based on the elevation of their residential area; elevation less than 10 meters, elevation between 10 meters and 50 meters, and elevation more than 50 meters. Also, to control for the effect of evacuation orders on individuals' decisions, we separated the dataset into three categories; individuals who received no evacuation order, individuals who received voluntary evacuation orders, and individuals who received mandatory evacuation orders. Evacuation rates for each group are presented in **Figure 10**. It can be seen that the elevation of residential area has a strong association with people's decision to evacuate. 36.59% of people who had not received any evacuation order but were living in low-lying residential areas decided to leave their home, while this rate was 28.43% for those in areas with elevation more than 50 meters. This observation is in accordance with the fact that people who live in areas with lower elevation are more concerned with the safety of their region during a hurricane.

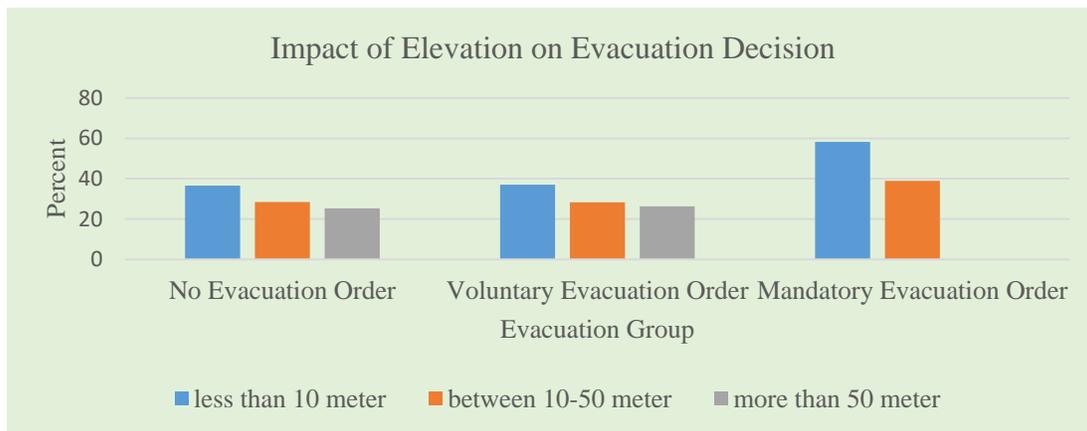

**Figure 10. Elevation impacts on evacuation decision**

## 6.  STATISTICAL MODEL

After constructing the evacuation pattern and conducting overall validation against existing polls and surveys, we investigated the statistical linkage between mobility patterns of individuals and their evacuation decisions. The evacuation decision has been well studied in the literature and its importance and implications for agencies have been highlighted. Previous studies indicated the importance of socio-demographic variables such as age, income, and race as well as evacuation orders and the perceived worries and concerns in evacuation decisions. In this paper, in addition to those metrics, we explored the importance of individuals' mobility behavior in their decision. We calculated individual level mobility measures such as the daily number of trips and the convex hull of each active device during August and incorporate these metrics to our logistic regression model to examine whether such observed mobility metrics can improve the evacuation decision model's accuracy. **Table 3** summarizes the list of variables considered in our models. To develop our statistical model, we removed 3,937 devices from our dataset who have missing values in some of their socio-demographic attributes.

**Table 3. Data descriptive for evacuation decision choice model**

| Metric | Definition | | Descriptive Statistics | | | |
|---|---|---|---|---|---|---|
| Categorical Variable | | | Count | Percentage | | |
| Evacuation Decision | Evacuation decision | 0 = did not evacuate, | 507605 | 63.16 | | |
| | | 1 = evacuate | 296081 | 36.84 | | |
| Evacuation order | Evacuation order received | 0 = none | 565178 | 70.32 | | |
| | | 1 = voluntary, | 114038 | 14.19 | | |
| | | 2 = mandatory | 124470 | 15.49 | | |
| Continuous Variable | | | Min | Median | Max | SD |
| Elevation | Residential location elevation | | -1 | 6 | 102 | 13.86 |
| Median age | Median age of the residential census tract | | 11.9 | 41.4 | 83.3 | 9.71 |
| Median income | Median income of the residential census tract | | 8804 | 54279 | 2500001 | 22951 |
| Vehicle availability | Percentage of households with at least one vehicle in the census tract | | 28.4 | 96.1 | 100 | 5.82 |
| Race - white | Percentage of white population in the census tract | | 0 | 0.83 | 1 | 0.17 |
| Average number of trip | Average number of trips taken by the individual per day during August | | 1 | 5.5 | 51.4 | 3.82 |
| Average of convex hull area | Average daily convex hull area of individuals during August | | 0 | 48.57 | 57274.8 | 510.31 |

As no evacuation was the base choice in our decision variable, positive coefficients indicate that increase in variables' value increases the likelihood of evacuation, while a negative sign denotes a decrease in the likelihood of evacuation. The summary of results is presented in **Table 4**.

**Table 4. Logistic regression models' summary**

| Variable | Model#1 – logistic model without mobility behavior metrics | | Model#2 – logistic model with mobility behavior metrics | |
|---|---|---|---|---|
| | Estimated coefficient | p-value | Estimated coefficient | p-value |
| Intercept | 3.61E-01 | <0.001 *** | 4.45 E-01 | <0.001 *** |
| Evacuation order | 4.06 E-01 | <0.001 *** | 4.08 E-01 | <0.001 *** |
| Elevation | -8.60 E-05 | <0.001 *** | -8.55 E-05 | <0.001 *** |
| Median age | 8.48 E-03 | <0.001 *** | 8.65 E-03 | <0.001 *** |
| Median income | 3.62 E-08 | 0.766 | 2.68 E-07 | 0.028 * |
| Vehicle availability | -1.57 E-02 | <0.001 *** | -1.88 E-02 | <0.001 *** |
| Race - white | 2.59 E-01 | <0.001 *** | 2.44 E-01 | <0.001 *** |
| Average number of trip | - | - | 1.03E-02 | <0.001 *** |
| Average of convex hull area | - | - | 4.28E-04 | <0.001 *** |
| Number of observation | 803686 | | 803686 | |
| Log Likelihood | -516912.5 (df=7) | | -513806.2 (df=9) | |
| AIC | 1033839 | | 1027630 | |
| McFadden R2 | 0.025 | | 0.031 | |
| Models Comparison | P-value (Chi) = <0.001 *** | | | |

We developed two logistic regression models. Model#1 only includes socio-demographic information, elevation of residential location, and evacuation order attributes while model#2 utilized mobility behavior metrics in addition to all variables in model#1. In both models, the sign of coefficients for common variables was in line and consistent with previous studies except for the vehicle availability metric. Higher vehicle availability was expected to increase the tendency to evacuate but in our model, the coefficient was estimated negative. One possible reason for this observation might be due to the low variation in this metric (the first quantile of vehicle availability was 92.6% and the median was 96.1%). As expected, both mobility metrics were statistically significant in model#2 and they improved the overall accuracy of the model significantly. The estimated sign of the coefficients was positive which shows that individuals with more trips per day and larger spatial trajectory area footprint are more likely to evacuate their residential location during a disaster.

# 7. CONCLUSION

The intensity and the frequency of weather-related disasters are expected to increase due to climate change, increase in sea surface temperature, and other related causes [42,43]. In order to be prepared, it is crucial for the state and government agencies to understand individuals' behavior before, during, and after a disaster. Most of the research in the literature studied individuals' behavior on these extreme events based on post-disaster surveys. In addition to small sample size, these surveys are typically prone to several biases, such as observer effect bias and imperfect recall of the evolution of the evacuation process. In this study, we constructed several aspects of evacuation patterns by analyzing anonymized individuals' traces using the mobile phone LBS data.

To study the evacuation pattern, we analyzed two months of mobile LBS data for more than 2 million users. We were able to capture the evacuation behavior of 807,623 anonymized individuals by employing our proposed framework on more than 11 billion location sightings. Our study showed that type of evacuation order has a strong impact on individuals' evacuation decisions. Results showed that 57.92% of individuals who received mandatory evacuation orders left their homes, while this ratio was 32.98% and 35.68% for smartphone users who received no evacuation order and voluntary evacuation order, respectively.

Irma made its landfall on September 10. The Departure date and reentry date analysis conducted in this paper demonstrated that the majority of the evacuees left their residences in the last three days leading to the hurricane landfall, with the peak of evacuation observed on September 9 when 26.27% of evacuees departed their home. However, the returning process was distributed more evenly among days after the landfall. We also empirically examined the effect of the evacuation order issue date on individuals' departure date decisions. It was shown that late evacuation orders (ones that issued on September 9 and September 10) did not have strong influence on individuals' departure decision; while for the regions that received evacuation orders earlier (from September 6 to September 8) an increase was observed in evacuation rate the day after the evacuation order was issued. These findings highlight the importance of issuing evacuation orders at least two days before hurricane landfall.

The evacuation distance distribution revealed that the majority of people either selected to shelter in the vicinity of their residential area or decided to go to farther destinations rather than to choose destinations within 20 to 100 miles. We also showed that elevation of the residential area had strong effect on individuals' evacuation decisions. People living in low-lying regions showed a higher evacuation rate in comparison to people living in mid and high elevation regions after controlling for the evacuation order type.

By developing an evacuation decision choice model, we showed that the observed mobility pattern of individuals can play a significant role in improving the accuracy of evacuation decision models. Having access to historical mobile device location data provides unique information to the agencies and decision-makers to have a better understanding and estimate of the evacuation evolution in their region.

Although analyzing the behavior of smartphone users provides a unique opportunity to observe the actual behavior of millions of individuals, several limitations exist. While the sample size of the mobile traces data is enormous, it should still be noted that these type of data have their own biases. The other limitation is the fact that post-disaster surveys usually provide a rich set of socio-demographic information of responders while in mobile LBS data there is no such information provided. However, recently many studies have been trying to impute missing information such as users' socio-demographics. Employing these developed approaches would be helpful to add new dimensions to our future analysis.

## ACKNOWLEDGMENTS

This research was partly supported by the Maryland Transportation Institute. We are also grateful to Cuebiq for providing us access to their data. Any opinions, findings, and conclusions or recommendations expressed are those of the authors and do not necessarily reflect the views of the sponsors.

## AUTHOR CONTRIBUTIONS

The authors confirm contribution to as follows: study conception and design: AD, VF, and LZ; data collection and processing: AD, SG, and HY; analysis and interpretation of results: AD, HY, VF, and LZ; draft manuscript preparation: AD, SG, VF, HY, and LZ. All authors reviewed the results and approved the final version of the manuscript.

## REFERENCES

1. Office for coastal management NOaAA. <**https://coast.noaa.gov/states/fast-facts/hurricane-costs.html**>.

2. Wong S, Shaheen S, Walker J. Understanding Evacuee Behavior: A Case Study of Hurricane Irma. 2018.

3. *Tropical Cyclone Report: Hurricane Irma.* National Hurricane Center.

4. Hasan S, Ukkusuri S, Gladwin H, Murray-Tuite P. Behavioral model to understand household-level hurricane evacuation decision making. *Journal of Transportation Engineering.* 2010;137(5): 341-348.

5. Smith SK, McCarty C. Fleeing the storm (s): An examination of evacuation behavior during Florida's 2004 hurricane season. *Demography.* 2009;46(1): 127-145.

6. Robinson RM, Foytik P, Jordan C. *Review and Analysis of User Inputs to Online Evacuation Modeling Tool.* 2017.

7. McCarney R, Warner J, Iliffe S, Van Haselen R, Griffin M, Fisher P. The Hawthorne Effect: a randomised, controlled trial. *BMC medical research methodology.* 2007;7(1): 30.

8. Groves RM. *Survey errors and survey costs.* Vol 536: John Wiley & Sons; 2004.

9. Furnham A. Response bias, social desirability and dissimulation. *Personality and individual differences.* 1986;7(3): 385-400.

10. Wang Q, Taylor JE. Quantifying human mobility perturbation and resilience in Hurricane Sandy. *PLoS one.* 2014;9(11): e112608.

11. Roy KC, Hasan S. *Modeling the dynamics of hurricane evacuation decisions from real-time Twitter data.* 2019.

12. Yabe T, Ukkusuri SV. Effects of income inequality on evacuation, reentry and segregation after disasters. *Transportation Research Part D: Transport and Environment.* 2020: 102260.

13. Yabe T, Tsubouchi K, Fujiwara N, Sekimoto Y, Ukkusuri SV. Understanding post-disaster population recovery patterns. *Journal of the Royal Society Interface.* 2020;17(163): 20190532.

14. Younes H, Darzi A, Zhang L. How effective are evacuation orders? An analysis of decision making among vulnerable populations in Florida during Hurricane Irma.: SocArXiv; 2021.

15. Huang S-K, Lindell MK, Prater CS. Who leaves and who stays? A review and statistical meta-analysis of hurricane evacuation studies. *Environment and Behavior.* 2016;48(8): 991-1029.

16. Murray-Tuite P, Wolshon B. Evacuation transportation modeling: An overview of research, development, and practice. *Transportation Research Part C: Emerging Technologies.* 2013;27: 25-45.

17. Wolshon PB. *Transportation's role in emergency evacuation and reentry.* Vol 392: Transportation Research Board; 2009.

18. Collier J, Balakrishnan S, Zhang Z. *From Hurricane Katrina to Hurricane Harvey: Actions, Issues, and Lessons Learned in Transportation and Logistics Efforts for Emergency Response.* 2019.

19. Yin W, Murray-Tuite P, Ukkusuri SV, Gladwin H. An agent-based modeling system for travel demand simulation for hurricane evacuation. *Transportation research part C: emerging technologies.* 2014;42: 44-59.

20. Brown C, White W, van Slyke C, Benson JD. Development of a strategic hurricane evacuation–dynamic traffic assignment model for the Houston, Texas, Region. *Transportation Research Record.* 2009;2137(1): 46-53.

21. Wang H, Mostafizi A, Cramer LA, Cox D, Park H. An agent-based model of a multimodal near-field tsunami evacuation: Decision-making and life safety. *Transportation Research Part C: Emerging Technologies.* 2016;64: 86-100.

22. Feng K, Lin N. Simulation of Hurricane Irma Evacuation Process. 2019.

23. Mostafizi A, Wang H, Dong S. Understanding the Multimodal Evacuation Behavior for a Near-Field Tsunami. *Transportation Research Record.* 2019: 0361198119837511.

24. Robinson RM, Collins AJ, Jordan CA, Foytik P, Khattak AJ. Modeling the impact of traffic incidents during hurricane evacuations using a large scale microsimulation. *International journal of disaster risk reduction.* 2018;31: 1159-1165.

25. Zhang Z, Wolshon B, Herrera N, Parr S. Assessment of post-disaster reentry traffic in megaregions

using agent-based simulation. *Transportation Research Part D: Transport and Environment.* 2019;73: 307-317.

26. Wu H-C, Lindell MK, Prater CS. Logistics of hurricane evacuation in Hurricanes Katrina and Rita. *Transportation research part F: traffic psychology and behaviour.* 2012;15(4): 445-461.

27. Liu S, Murray‐Tuite P, Schweitzer L. Incorporating Household Gathering and Mode Decisions in Large‐Scale No‐Notice Evacuation Modeling. *Computer‐Aided Civil and Infrastructure Engineering.* 2014;29(2): 107-122.

28. Yang H, Morgul EF, Ozbay K, Xie K. Modeling evacuation behavior under hurricane conditions. *Transportation research record.* 2016;2599(1): 63-69.

29. Kontou E, Murray-Tuite P, Wernstedt K. Duration of commute travel changes in the aftermath of Hurricane Sandy using accelerated failure time modeling. *Transportation Research Part A: Policy and Practice.* 2017;100: 170-181.

30. Wong S, Walker J, Shaheen S. Assessing the Feasibility and Equity Impacts of the Sharing Economy in Evacuations: A Case Study of Hurricane Irma TRB 2019 Annual Meeting; 2019.

31. Kumar D, Ukkusuri SV. Utilizing geo-tagged tweets to understand evacuation dynamics during emergencies: A case study of Hurricane Sandy. Paper presented at: Companion Proceedings of the The Web Conference 20182018.

32. <**http://geodata.floridadisaster.org/datasets/555551c57224402bb4c2387868cc8c93_1**>.

33.          <**https://web.archive.org/web/20180125015640/https://www.flgov.com/2017/09/09/gov-scott-issues-updates-on-hurricane-irma-preparedness-10/**>. July 2019.

34. Alexander L, Jiang S, Murga M, González MC. Origin–destination trips by purpose and time of day inferred from mobile phone data. *Transportation research part c: emerging technologies.* 2015;58: 240-250.

35. Wang F, Chen C. On data processing required to derive mobility patterns from passively-generated mobile phone data. *Transportation Research Part C: Emerging Technologies.* 2018;87: 58-74.

36. Zahedi S, Shafahi Y. Estimating activity patterns using spatio-temporal data of cell phone networks. *International Journal of Urban Sciences.* 2018;22(2): 162-179.

37. Ester M, Kriegel H-P, Sander J, Xu X. A density-based algorithm for discovering clusters in large spatial databases with noise. Paper presented at: Kdd1996.

38. Zhang L, Ghader S, Darzi A. Data Analytics and Modeling Methods for Tracking and Predicting Origin-Destination Travel Trends Based on Mobile Device Data. *Federal Highway Administration Exploratory Advanced Research Program.* 2020.

39. Williams NE, Thomas T, Dunbar M, Eagle N, Dobra A. Measurement of human mobility using cell phone data: developing big data for demographic science. Paper presented at: Population Association of America Annual Meeting2013.

40. Csáji BC, Browet A, Traag VA, et al. Exploring the mobility of mobile phone users. *Physica A: statistical mechanics and its applications.* 2013;392(6): 1459-1473.

41.        Hurricane      Irma.       <**https://media.news4jax.com/document_dev/2017/10/26/Mason-Dixon%20Hurricane%20poll_1509043928726_10861977_ver1.0.pdf**>. Accessed October 25 2017.

42. Elsner JB, Kossin JP, Jagger TH. The increasing intensity of the strongest tropical cyclones. *Nature.* 2008;455(7209): 92.

43. Webster PJ, Holland GJ, Curry JA, Chang H-R. Changes in tropical cyclone number, duration, and intensity in a warming environment. *Science.* 2005;309(5742): 1844-1846.